\pdfoutput=1

\documentclass[11pt]{article}

\usepackage[]{acl}

\usepackage{times}
\usepackage{latexsym}
\usepackage{graphicx}
\usepackage{hyperref}
\usepackage{tabularx}
\usepackage{url}
\usepackage{float}
\usepackage{booktabs}
\usepackage{multicol, multirow}
\usepackage{wrapfig}
\usepackage{subfigure}
\usepackage{enumitem}
\usepackage{color,soul}
\usepackage{longtable}
\usepackage{diagbox}
\usepackage{tabularray}
\usepackage{todonotes}

\usepackage[T1]{fontenc}

\usepackage[utf8]{inputenc}

\usepackage{microtype}

\usepackage{inconsolata}
\usepackage{amsmath}
\usepackage{amssymb}
\bibpunct[, ]{(}{)}{;}{a}{,}{,}

\title{Generating Zero-shot Abstractive Explanations for Rumour Verification}

\author{Iman Munire Bilal$^{1,2}$, Preslav Nakov$^{2}$, Rob Procter$^{1,3}$, Maria Liakata$^{1,3,4}$ \\
$^1$ University of Warwick \\
$^2$ Mohammed Bin Zayed University of Artificial Intelligence \\
$^3$ The Alan Turing Institute\\
$^4$ Queen Mary University of London\\
   \texttt{$\{$iman.bilal|rob.procter$\}$@warwick.ac.uk}\\
  \texttt{preslav.nakov@mbzuai.ac.ae m.liakata@qmul.ac.uk}\\}

\begin{document}
\maketitle

\begin{abstract}
The task of rumour verification in social media concerns assessing the veracity of a claim on the basis of conversation threads that result from it. While previous work has focused on predicting a veracity label, here we reformulate the task to generate model-centric free-text explanations of a rumour's veracity. The approach is model agnostic in that it generalises to any model. Here we propose a novel GNN-based rumour verification model. We follow a zero-shot approach by first applying post-hoc explainability methods to score the most important posts within a thread and then we use these posts to generate informative explanations using opinion-guided summarisation. To evaluate the informativeness of the explanatory summaries, we exploit the few-shot learning capabilities of a large language model (LLM). Our experiments show that LLMs can have similar agreement to humans in evaluating summaries. Importantly, we show explanatory abstractive summaries are more informative and better reflect the predicted rumour veracity than just using the highest ranking posts in the thread.
\footnote{Our code and data is found at \url{https://github.com/bilaliman/RV_explainability}.}
\end{abstract}

\section{Introduction}

Evaluating misinformation on social media is a challenging task that requires many steps \citep{pheme_original}: detection of rumourous claims, identification of stance towards a rumour, and finally assessing rumour veracity. In particular, misinformation may not be immediately verifiable using reliable sources of information such as news articles since they might not have been available at the time a rumour has emerged.
For the past eight years, researchers have focused on the task of automating the process of rumour verification in terms of assigning a label of \emph{true}, \emph{false}, or \emph{unverified} \citep{pheme_original,derczynski-etal-2017-semeval}. However, recent work has shown that while fact-checkers agree with the urgent need for computational tools for content verification, the output of the latter can only be trusted if it is accompanied by explanations 
\citep{procter2023observations}.

Thus, in this paper, we move away from black-box classifiers of rumour veracity to generating explanations written in natural language (free-text) for why, given some evidence, a statement can be assigned a particular veracity status. This has real-world applicability particularly in rapidly evolving situations such as natural disasters or terror attacks \citep{riots}, where the explanation for an automated veracity decision is crucial \citep{lipton}.
To this effect, we use a popular benchmark, the PHEME \citep{pheme_original} dataset, to train a rumour verifier and employ the  conversation threads that form its input to generate model-centric explanation summaries of the model's assessments.

\citet{atanasova-etal-2020-generating-fact}, \citet{kotonya-toni-2020-explainable-automated} and \citet{Stammbach2020eFEVEREA} were the first to introduce explanation summaries for fact-checking across different datasets.
\citet{kotonya-toni-2020-explainable-automated} provided a framework for creating abstractive summaries that justify the true veracity of the claim in the PUBHealth dataset, similarly to \citet{Stammbach2020eFEVEREA} who augment the FEVER~\citep{thorne-etal-2018-fact} dataset with a corpus of explanations. \citet{atanasova-etal-2020-generating-fact} proposed a jointly trained system that produces veracity predictions as well as explanations in the form of extracted evidence from ruling comments on the LIAR-PLUS dataset \cite{alhindi-etal-2018-evidence}. The approach in \cite{kotonya-toni-2020-explainable-automated} results in explanatory summaries that are, however, not faithful to the model, while \citet{atanasova-etal-2020-generating-fact} requires a supervised approach.
Our goal is to create a novel zero-shot method for abstractive explanations that explain the rumour verification model's predictions. 
We make the following contributions:
\begin{itemize}[nosep,leftmargin=*]
\item We introduce a zero-shot framework for generating abstractive explanations using opinion-guided summarisation for the task of rumour verification. To the best of our knowledge, this is the first time free-text explanations are introduced for this task.
\item We investigate the benefits of using a gradient-based algorithm and a game theoretical algorithm to provide explainability.
\item While our explanation generation method is generalisable to any verification model, we introduce a novel graph-based hierarchical approach.
\item We evaluate the informativeness of several explanation baselines, including model-independent and model-dependent ones 
stemming from the highest scoring posts by providing them as input to a few-shot trained large language model. Our results show that our proposed abstractive model-centric explanations are more informative in $77\%$ of the cases as opposed to $49\%$ for all other baselines.

\item We provide both human and LLM-based evaluation of the generated explanations, showing that LLMs achieve sufficient agreement with humans, thus allowing scaling of the evaluation of the explanatory summaries in absence of gold-truth explanations.
\end{itemize}
\begin{figure*}[!ht]
  \includegraphics[width=\textwidth]{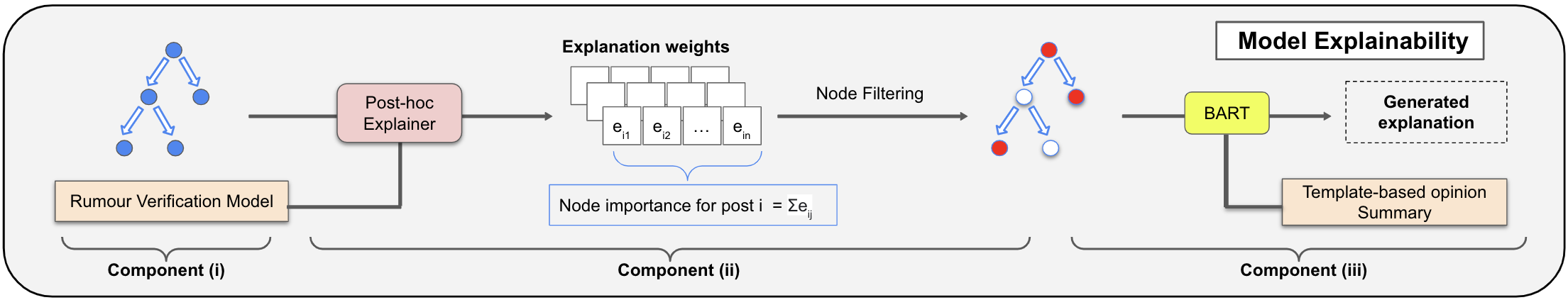}
  \caption{Framework of our proposed approach to obtain faithful generated explanations for the rumour verification model. It explains the process of explanation generation, where the weights from a model are passed through an explainer algorithm to identify important input nodes, which are then filtered and used in abstractive summarisation.}
\label{overview_fig}
\end{figure*}

\section{Related Work}

\paragraph{Explainable Fact Checking} 
Following the example of fact-checking organisations (e.g., Snopes, Full Fact, Politifact), which provide journalist-written justifications to determine the truthfulness of claims, recent datasets augmented with free-text explanations have been constructed: LIAR-PLUS~\citep{alhindi-etal-2018-evidence}, PubHealth~\citep{kotonya-toni-2020-explainable-automated}, AVeriTeC~\citep{schlichtkrull2023averitec}. A wide range of explainable outputs and methods have been proposed: theorem proofs~\citep{krishna-etal-2022-proofver}, knowledge graphs~\citep{ahmadi2019explainable}, question-answer decompositions~\citep{boissonnet-etal-2022-explainable,chen-etal-2022-generating}, reasoning programs~\citep{pan-etal-2023-fact}, deployable evidence-based tools~\citep{faxplainac} and summarisation \citep{atanasova-etal-2020-generating-fact,kotonya-etal-2021-graph,Stammbach2020eFEVEREA,kazemi-etal-2021-extractive,jolly}. We adopt summarisation as our generation strategy as it fluently aggregates evidence from multiple inputs and has been proven effective in similar works which we discuss next.

\paragraph{Explainability as Summarisation} 
\citet{atanasova-etal-2020-generating-fact} and \citet{kotonya-toni-2020-explainable-automated} leveraged large-scale datasets annotated with gold justifications to generate supervised explanations for fact-checking, while \citet{Stammbach2020eFEVEREA} used few-shot learning on GPT-3 to create the e-FEVER dataset of explanations. Similar to \citep{Stammbach2020eFEVEREA}, \citet{kazemi-etal-2021-extractive} also leveraged a GPT-based model (GPT-2) to generate abstractive explanations, but found that that their extractive baseline, Biased TextRank, outperformed GPT-2 on the LIAR-PLUS dataset \citep{alhindi-etal-2018-evidence}. \citet{jolly} warn that the output of extractive explainers lacks fluency and sentential coherence, which motivated their work on unsupervised post-editing using the explanations produced by \citet{atanasova-etal-2020-generating-fact}.
Our approach is different from the above as we derive our summaries from microblog content (as opposed to news articles as done by \citet{atanasova-etal-2020-generating-fact, Stammbach2020eFEVEREA, kazemi-etal-2021-extractive, jolly}, and only use the subset of posts relevant to the model's decision to inform the summary (rather than summarising the whole input as in \citep{kotonya-toni-2020-explainable-automated, kazemi-etal-2021-extractive}. Moreover, we rely on a zero-shot generation approach without gold explanations, contrary to \citep{atanasova-etal-2020-generating-fact, kotonya-toni-2020-explainable-automated}.

\paragraph{LLMs as evaluators}
Having generated explanatory summaries the question arises as to how to evaluate them at scale. LLMs have been employed as knowledge bases for fact-checking~\citep{lee-etal-2020-language,pan-etal-2023-fact}, as explanation generators for assessing a claim's veracity~\citep{Stammbach2020eFEVEREA,kazemi-etal-2021-extractive} and, as of recently, as evaluators in generation tasks. Most works focused on assessing the capability of LLM-based evaluation on summarisation tasks, either on long documents~\citep{wu2023long} or for low-resource languages~\citep{hada2023large}. While there is work focusing on reducing positional bias~\citep{wang2023large} and costs incurred~\citep{wu2023long} for using LLM-based evaluators, our evaluation is most similar to \citet{liu2023geval, shen2023large,chiang2023large}, who study the extent of LLM-human agreement in evaluations of fine-grained dimensions, such as fluency or consistency. We believe we are the first to use an LLM-powered evaluation to assess the informativeness and faithfulness of explanations for verifying a claim.

\section{Methodology}

Our methodological approach (Figure~\ref{overview_fig}) consists of three individual components: \emph{i})~training a rumour verification model; \emph{ii})~using a post-hoc explainability algorithm; \emph{iii})~generating summary-explanations. The approach to explanation generation is zero-shot and model-agnostic. 

We demonstrate our approach on PHEME~\citep{pheme_original}, a widely used benchmark dataset for classifying social media rumours into either unverified, true or false. It contains conversation threads that cover 5 real-world events such as the Charlie Hebdo attack and the Germanwings plane crash. We adopt the same leave-one-out testing as previous works \citep{dougrez-lewis-etal-2022-phemeplus} which favours real-world applicability as the model is tested on new events not included in the test data.

\paragraph{Task Formulation} For a model trained on rumour verification $\mathcal{M}$, an attribution-based explanation method $\mathcal{E}$, and a rumourous conversation thread consisting of posts $\mathcal{T} = \{p_1,...p_l \}$ with embeddings $\{x_1,...x_l \}\subset\mathbb{R}^n$, we define the post importance as a function $f_{(\mathcal{M},\mathcal{E})}:\mathcal{T}\rightarrow \mathbb{R}$.

\begin{equation} \label{eq:1} f_{(\mathcal{M},\mathcal{E})}(p_i) = \sum_{j=1}^{n} \mathcal{E}(\mathcal{M},x_i)_j = \sum_{j=1}^{n}e_{ij} \end{equation}

\noindent where $e_i \in \mathbb{R}^n$ is the attribution vector for embedding $x_i$ of post $p_i$ such that each value $e_{ij}$ corresponds to the weight of feature $x_{ij}$ assigned by the explainer algorithm $\mathcal{E}$.

The summary is generated from the subset of posts that are most important for the model prediction, i.e., $\mathcal{I}=\{p_i \mid f_{(\mathcal{M},\mathcal{E})}(p_i)>0 \}$. Note a thread will contain posts that agree with the prediction (positive importance scores) and posts that disagree (negative importance scores).

\subsection{Rumour Verification Model}
\label{3.1}

\begin{figure*}[!ht]
  \includegraphics[width=\textwidth]{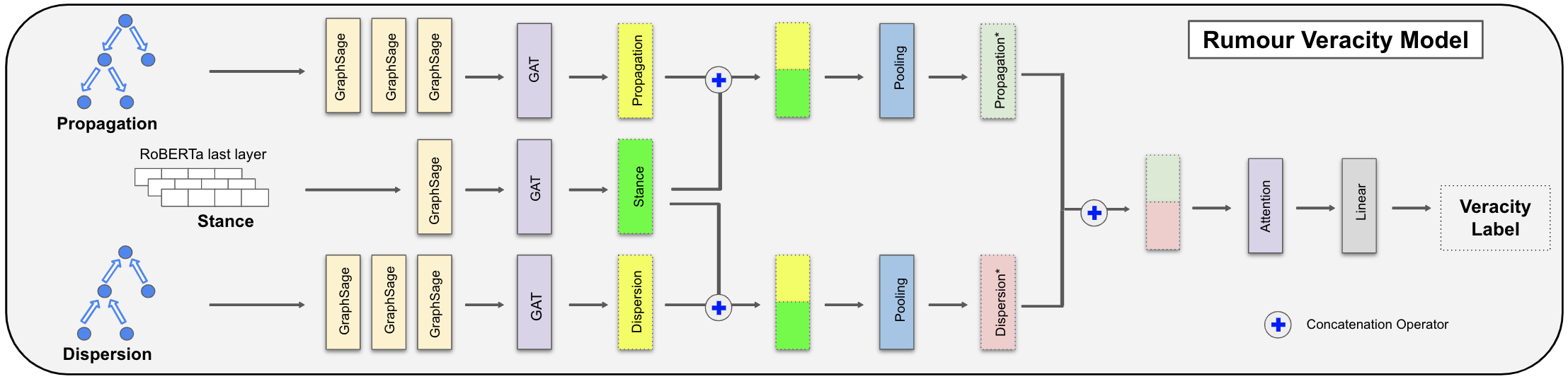}
  \caption{Architecture of our rumour verification model enhanced with structure-aware and stance-aware components based on graph neural networks. In the diagram, Propagation/Dispersion/Dispersion represent the outputs of each respective component, while Propagation*/Dispersion* represent the stance-enriched outputs of these.}
\label{architecture}
\end{figure*}

Our explanation generation method is applicable to any rumour verification model, but here we chose an approach based on graph neural networks (See Figure \ref{architecture}), which caters for a flexible information structure combining information in the conversation thread with information about stance. This is the first time a GNN-based model enriched with stance has been proposed for PHEME.

\begin{table}[!ht]
    \centering
    \scriptsize
    \begin{tabular}{@{}m{2.5cm}cccccc@{}}
    \toprule
        & F& C& O & G& S& F1\\
        \hline
        Our model w/o stance & .228& .267&.300&.333&.293&.405  \\
        \midrule
        Our model with stance &.208&.341&.313&.403&.358&.434 \\
        \midrule
        SAVED \citep{dougrez-lewis-etal-2021-learning} & .372& .351&  .304& .281& .332&.434 \\
        \bottomrule
     
    \end{tabular}
    \caption{PHEME results for each fold and overall reported as macro-averaged F1 scores. The fold abbreviations stand for Ferguson, Charlie Hebdo, Ottawa Shooting, Germanwings Crash and Sydney Siege}.
    \label{tab:pheme_performance}
\end{table}

\paragraph{Structure-Aware Model}
Structure-aware models such as tree-based and graph-based are among the best performing for rumour verification \citep{kochkina-etal-2018-one, Bian_Xiao_Xu_Zhao_Huang_Rong_Huang_2020,KOCHKINA2023103116}, given that the task heavily relies on user interactions for determining veracity. Our approach  models the conversation thread as a graph, where interactions between posts manifest as propagation (top-down) and dispersion (bottom-up) flows similar to \citet{Bian_Xiao_Xu_Zhao_Huang_Rong_Huang_2020}. 
The architecture contains GraphSage \citep{graphsage} layers, proven to yield meaningful node representations, followed by GAT \citep{velivckovic2017graph} layers, which are shown to improve performance in similar tasks \citep{kotonya-etal-2021-graph, gat_application, jia2022early}. Sentence Transformers embeddings \citep{reimers-2019-sentence-bert} are used to initialise the node representations in the graphs. The propagation and dispersion component outputs are each concatenated with the output of a stance component and pooled, resulting in another concatenated representation to which a final multi-head attention layer \citep{NIPS2017_3f5ee243} is applied. 

\paragraph{Stance-Aware Component}
Stance detection is closely linked to misinformation detection \citep{hardalov-etal-2022-survey} with previous work having shown that a joint approach improves rumour verification \citep{pheme_original, derczynski-etal-2017-semeval, gorrell-etal-2019-semeval, yu-etal-2020-coupled, dougrez-lewis-etal-2021-learning}. 
As such our model includes a stance component unlike the GNN by \citet{Bian_Xiao_Xu_Zhao_Huang_Rong_Huang_2020}. Since only a small portion of the PHEME dataset is annotated with gold stance labels for the RumourEval competition 
\citep{derczynski-etal-2017-semeval}, we generate silver labels for the whole corpus. In particular, we train a RoBERTa model \citep{roberta} for stance classification and extract the embeddings from the last hidden layer to augment the rumour verification task with stance information. See Appendix \ref{sec:appendixD} for experimental setup.

\paragraph{Performance of Rumour Verification Baselines}
We include the performance of our proposed baselines, the structure-aware model and its stance-aware version, in Table \ref{tab:pheme_performance}. 

As expected, integrating stance knowledge into the model boosts performance by almost 3 F1-points overall with improved scores across the majority of folds; we hypothesise performance does not improve for the Ferguson fold due to its severe label imbalance skewed towards unverified rumours. Moreover, we observe that the model enhanced with the stance-aware component achieves competitive results and is comparable to the current state-of-the-art model on the PHEME dataset, the SAVED model by~\citet{dougrez-lewis-etal-2021-learning}. 

\subsection{Explaining the Model}
\label{3.2}

\subsubsection{Attribution Method} 
We experiment with two classes of attribution methods: gradient-based and game-theory-based.
For gradient-based methods, we choose Integrated Gradients (IG) \citep{integrated_gradients}. This is a local explainability algorithm that calculates attribution scores for each input unit by accumulating gradients along the interpolated path between a local point and a starting point with no information (baseline). IG was selected over other gradient-based saliency methods such as DeepLIFT \citep{shrikumar2017learning} as it has been shown to be more robust \citep{pruthi} when applied in classification tasks.
Shapley Values (SV) \citep{shapley_val} is the representative explainability method derived from game theory and has been used in many applications \citep{shap_fin,mosca-etal-2021-understanding,ekbal-2022-adversarial}. Its attribution scores are calculated as expected marginal contributions where each feature is viewed as a 'player' within a coalitional game setting.

Note that we focus on post-hoc methods instead of intrinsic ones, such as attention, in our architecture to keep the framework generalisable to other rumour verification models. 
Specifically, we use IG and SV\footnote{Used \emph{captum} package~\citep{kokhlikyan2020captum} for both.} as methods for $\mathcal{E}$ to calculate the post importance $f$ in Equation~\ref{eq:1}. This importance with respect to model prediction is then leveraged to sort the posts within the thread in descending order. 
We then construct subsets of important posts $\mathcal{I}_k \subset \mathcal{I}$ such that $|\mathcal{I}_k|=k \% |\mathcal{I}|$ with $\mathcal{I}_k$ representing the $k \%$ most important posts of the rumour thread, $k={25,50,100}$. These will be used as inputs for summarisation in the next stage to determine the trade-off between post importance and number of posts necessary to construct a viable explanation.

\subsubsection{Summarisation for Explanation}
We propose explanation baselines spanning different generation strategies: extractive 
vs abstractive, model-centric vs model-independent and in-domain vs out-of-domain.

\paragraph{Extractive Explanations}
\begin{itemize}[nosep,leftmargin=*]
    \item \textit{Important Response}: the response within the thread scored as most important by each attribution method. This is a model-dependent explanation.
    \item \textit{Similar Response}: the response within the thread most semantically similar to the source claim, as scored by SBERT \citep{reimers-2019-sentence-bert}. This model-independent baseline is inspired by \citep{russo-etal-2023-benchmarking}.
\end{itemize}

\paragraph{Abstractive explanations} have a dual purpose that fits the challenging set-up of our pipeline: they serve as a way to aggregate important parts of the thread, and also provide a fluent justification sourced from multiple views to a claim's veracity.
\begin{itemize}[nosep,leftmargin=*]
    \item \textit{Summary of $\mathcal{I}$}: We summarise the set $\mathcal{I}$ of important posts to obtain a model-centric explanation. We fine-tune BART \citep{lewis-etal-2020-bart} on the MOS corpus introduced by \citet{bilal-etal-2022-template} that addresses summarisation of topical groups of tweets by prioritising the majority opinion expressed. We hypothesise this template-guided\footnote{The template summary takes the form: \textit{Main Story} + \textit{Majority} Opinion expressed in the thread.} approach will satisfy the explanatory purpose since user opinion is an important indicator for assessing a claim's veracity in rumour verification \citep{hardalov-etal-2022-survey}. Similarly, we define explanations \textit{Summary of $\mathcal{I}_{25}$} \& \textit{Summary of $\mathcal{I}_{50}$}.

    \item\textit{Out-of-domain Summary}: We use the BART \citep{lewis-etal-2020-bart} pre-trained on the CNN/ Daily Mail \citep{nallapati-etal-2016-abstractive} dataset without any fine-tuning and summarise the entire thread. This yields a model-independent explanation.
\end{itemize}

\noindent We note that while supervised summarisation is used to inform our generation strategy, our resulting explanations never rely on gold explanations annotated for the downstream task of fact-checking. In fact, neither MOS~\citep{bilal-etal-2022-template} nor the CNN/Daily Mail~\citep{nallapati-etal-2016-abstractive} datasets were aimed for fact-checking and both focus on broad topics unrelated to the PHEME claims.

\section{Automatic Evaluation of Explanation Quality}
\label{Sec:3}

\begin{table}[!h]
    \centering
    \tiny
    \begin{tabular}{p{7.5cm}}
\toprule
\multirow{4}{7cm}{You will be shown a \textbf{Claim} and an \textbf{Explanation}. The veracity of the Claim can either be true, false or unverified. Choose an option from A to D that answers whether the Explanation can help confirm the veracity of the Claim.}\\ \\ \\ \\

\multirow{3}{7.5cm}{\textbf{A}: The Explanation confirms the information in the Claim is true. The Explanation will include evidence to prove the Claim or show users believing the Claim.}\\ \\ \\
\multirow{3}{7.5cm}{\textbf{B}: The Explanation confirms the information in the Claim is false. The Explanation will include evidence to disprove the Claim or show users denying the Claim.}\\ \\ \\
\multirow{3}{7.5cm}{\textbf{C}: The Explanation confirms the information in the Claim is unverified. The Explanation will state no evidence exists to prove or disprove the Claim or show users doubting the Claim.}\\ \\ \\
\multirow{4}{7.5cm}{\textbf{D}: The Explanation is irrelevant in confirming the veracity of the Claim. The Explanation will not include any mention of evidence and users will not address the veracity of the Claim.}\\ \\ \\ \\

\textbf{Claim}: \textcolor{blue}{$\{$claim$\}$}\\
\textbf{Explanation}: \textcolor{blue}{$\{$explanation$\}$}\\
\bottomrule
          
    \end{tabular}
    \caption{Example task instructions used in the prompt following a multiple-choice setting.}
    \label{tab:ex_prompt_guidelines}
\end{table}
\normalsize

As the PHEME dataset lacks gold standard explanations to compare against, we prioritise the extrinsic evaluation of the generated explanations with respect to their usefulness in downstream tasks. This is similar to work on explanatory fact-checking \citep{Stammbach2020eFEVEREA, krishna-etal-2022-proofver}.

In particular, we use the criterion of \textbf{informativeness} defined by \citet{atanasova-etal-2020-generating-fact} as the ability to deduce the veracity of a claim based on the explanation. If the provided explanation is not indicative of any veracity label (\emph{true}, \emph{false}, or \emph{unverified}), the explanation is considered uninformative. Otherwise, we compare the veracity suggested by the explanation to the prediction made by the model. This enables the evaluation of the explanation's fidelity to the model and is one of the main approaches to assess explanatory \textbf{faithfulness} in the research community \citep{jacovi-goldberg-2020-towards}.

We devise a novel evaluation strategy for capturing the informativeness of generated explanations based on LLMs. This is motivated by recent work demonstrating the effectiveness of LLM reasoning capability in various tasks \citep{kojima2022large, chen-2023-large}, including as a zero-shot evaluator for summarisation outputs \citep{liu2023geval,shen2023large, wang2023chatgpt}. We use OpenAI's \emph{gpt-3.5-turbo-0301}\footnote{Used GPT-3.5-turbo due to its lower running costs compared to GPT-4.}, hereinafter referred to as ChatGPT, which is a snapshot of the model from 1 March 2023 that will not receive updates – this should encourage the reproducibility of our evaluation. We follow a multiple-choice setting in the prompt similar to \citet{shen2023large}. Our initial experiments confirmed previous findings \citep{brownllm} that GPT reasoning can be improved by including a few annotated representative examples of the evaluation within its prompt (See Appendix \ref{sec:appendixA}). We experimented with several prompt designs varying in level of detail (no justification of answer, no examples) and found that the most exhaustive prompt yielded best results. The final task instructions used for the prompt are in Table~\ref{tab:ex_prompt_guidelines}. 

We ran a pilot study (See Appendix \ref{sec:appendixC}) to establish which temperature setting yields the most robust LLM evaluation. To account for any non-deterministic behaviour, the experiment was run three times. We find the results remain 100$\%$ consistent across runs for temperature 0. As this is in line with the settings used in similar works employing LLMs as evaluators \citep{shen2023large}, we also use this value for our experiment. Each request is sent independently via the Open AI API. Since using an LLM evaluator allows us to scale our evaluation \citep{chiang2023large}, we use all suitable PHEME threads\footnote{Suitable defined as at least ten posts and the majority are non-empty after URL and user mentions are removed.} (i.e. 1233 / 2107 threads) for testing. This set-up foregoes the costs necessary to obtain a diverse manually-annotated test set and offers more statistical power to the results as recommended by \citet{bowman-dahl-2021-will}.

\section{Results and Discussion}
\label{sec:5}

\noindent The results are shown in Table~\ref{tab:eval_info}.

\begin{table}[!ht]
    \centering
    \scriptsize
    
    \begin{tabular}{m{2.5cm}m{1.3cm}m{1.1cm}m{1.1cm}}

         & \textbf{Uninformative} & \textbf{Unfaithful} & \textbf{Faithful}\\
        \midrule
         &\multicolumn{3}{c}{Extractive Explanations} \\
        \midrule
         Important Response (IG) & 67.23 & \textbf{21.33} & 11.44\\
         Important Response (SV) & 65.29 & 22.30 & 12.41 \\
         Similar Response & 30.98 & 43.88 & 25.14 \\
         \midrule
         &\multicolumn{3}{c}{Abstractive Explanations} \\
        \midrule
        Summary of $\mathcal{I}_{25}$ (IG) & 23.68 & 46.55 & 29.76 \\
        Summary of $\mathcal{I}_{25}$ (SV) & 22.95 & 48.50 & 28.55 \\
        Summary of $\mathcal{I}_{50}$ (IG) & \textbf{22.11} & 46.47 & \textbf{30.41} \\
        Summary of $\mathcal{I}_{50}$ (SV) & 23.60 & 47.20 & 29.20 \\
        Summary of $\mathcal{I}$ (IG) & 24.90 & 48.58 & 26.52 \\
        Summary of $\mathcal{I}$ (SV) & 23.60 & 48.90 & 27.49 \\
        Out-of-domain Summary & 39.17 & 38.28 & 22.55 \\

    \end{tabular}
    \caption{Explanation evaluation wrt model prediction (\%). If the explanation cannot be used to infer a veracity label for the claim, it is \textbf{uninformative}. Otherwise, the explanation is \textbf{faithful} if its label coincides with the prediction and \textbf{unfaithful} if not. Best scores are in bold.}
    \label{tab:eval_info}
\end{table}
\normalsize

\paragraph{Model-centric vs Model-independent} We note that the explanations \textit{Out-of-domain Summary} and \textit{Similar Response} are independent of the rumour verification model built in section \ref{3.1} as they are not produced by any of the post-hoc algorithms. Hence, while these are not expected to be faithful, we analyse how they compare in informativeness to the other model-centric explanations. We find that abstractive explanations (\textit{Summaries of $\mathcal{I}_{25}$, $\mathcal{I}_{50}$, $\mathcal{I}$}) informed by the rumour verifier are the most informative of all. Thus, summarising a selection of important posts learned during the rumour verification process yields a better explanation than relying on individual replies or summarising the whole thread.

\paragraph{Integrated Gradients vs Shapley Values} The summaries generated via IG achieve better scores than the SV ones in both informativeness and faithfulness. While SV initially provides a better \textit{Important Response}, it fails to detect other important posts within the thread as suggested by the scores for $I_{25}$ and $I_{50}$. Moreover, the time complexity for the SV algorithm is exponential as its sampling strategy increases proportionately with the number of perturbed input permutations. We note the average computation time for both algorithms to assess a thread of 15 posts: 0.5s for IG and 2011s for SV. This makes IG a more suitable algorithm with respect to both performance and running time.

\paragraph{Extractive Explanation} 
The best extractive baseline is the \textit{Similar Response}, which selects the closest semantic match from the thread to the claim. Followed by are model-centric baselines \textit{Important Response} for both IG and SV, lagging behind by a large margin.  
We investigate the reason behind this performance by checking the stance labels of the corresponding posts. Using the labelled data from \citet{derczynski-etal-2017-semeval}, we train a binary RoBERTa to identify comments and non-comments\footnote{The original task is a 4-way classification of posts into one of the stance labels: \emph{support}, \emph{deny}, \emph{query}, or \emph{comment}. This is simplified by aggregating the first three labels into one.} where a comment is defined as a post that is unrelated or does not contribute to a rumour's veracity. We find that 64\% of posts corresponding to \textit{Important Response} labelled as uninformative are also classified as comments, much higher than 47\% for \textit{Similar Response}. This explains why semantic similarity can uncover a more relevant explanation than the \textit{Important Response} alone. Still, this method suffers from 'echoing' the claim \footnote{The majority of informative \textit{Similar Responses} are classified as supporting the claim.}, which risks missing out on other important information found in the thread (see Table ~\ref{Explanation_example}).
    \begin{table}
    \centering
    \tiny
    \begin{tabular}{p{7.5cm}}
        \toprule
        \textbf{Claim}\\
        \multirow{2}{7.5cm}{Update from Ottawa: Cdn soldier dies from shooting -Parliamentary guard wounded Parliament Hill still in lockdown URL}
        \\ \\

        \midrule
        \textbf{Prediction}: \textcolor{red}{Unverified} \\
        \midrule
        \textbf{Explanation Summaries}\\ \\
    
        \multirow{2}{7.5cm}{\textbf{Important Response}: @TorontoStar Ok, time to take it to the *** muslims. Look out Allah, here comes the revenge. ***. (\textit{Uninformative})}\\ \\

        \multirow{2}{7.5cm}{\textbf{Similar Response}: $\#$AttackinOttawa @TorontoStar: Update Cdn soldier dies from shooting -Parliamentary guard wounded Parliament Hill still in lockdown URL (\textit{True})}\\ \\
        
        \multirow{4}{7.5cm}{\textbf{Summary of $\mathcal{I}_{25}$ (IG)}: Soldier dies from shooting in Ottawa and Parliament Hill is in lockdown. \textcolor{red}{The majority think the media is wrong to report that Parliament Hill was in lockdown} and that the lockdown was a ploy to target Muslims. (\textit{False})}\\ \\ \\ \\

        \multirow{3}{7.5cm}{\textbf{Summary of $\mathcal{I}_{50}$ (IG)}: Cdn soldier dies from shooting dead in Ottawa. \textcolor{red}{The majority are sceptical about the news of the shooting and some are questioning where the confirmation is coming from.} (\textit{Unverified})}\\ \\ \\ 

        \multirow{4}{7.5cm}{\textbf{Summary of $\mathcal{I}$ (IG)}: Cdn soldier dies from shooting in Ottawa and Parliament Hill is in lockdown. Most users ask where the news of the gunman is and are wondering who is responsible for his death. Many of the responses use humour and irony, such as: \textcolor{red}{'I don't think the soldier is dead'}. (\textit{Unverified})}\\ \\ \\ \\

        \multirow{3}{7.5cm}{\textbf{Out-of-domain Summary}: Update from Ottawa: -Cdn soldier dies from shooting -Parliamentary guard wounded. \textcolor{red}{It looks like confirmations are coming in now. I don't think the soldier is dead.} (\textit{Unverified})}\\ \\ \\

        \bottomrule
    \end{tabular}
    \caption{Example explanation summaries. 
    Manually-annotated \textcolor{red}{red} highlights explain the model prediction for the given claim. ChatGPT evaluations are in ().}
    \label{Explanation_example}
\end{table}
\normalsize

\paragraph{Abstractive Explanation} 
The abstractive explanations are shown to be considerably more informative than most extractive baselines. They have the advantage of aggregating useful information that appears later in the conversation. For instance, the abstractive explanations in Table~\ref{Explanation_example} indicate posters' doubt and requests for more details. Furthermore, using an opinion-driven summariser is better for constructing a more informative summary-explanation than other options (See Sec. \ref{3.2}). We have also investigated the degree of information decay in relation to the number of posts used for summary construction in model-centric explanations. 
In Table~\ref{tab:eval_info}, the summary based on the first half of important posts ($\mathcal{I}_{50}$) yields the most informative and faithful explanation for both algorithms, closely followed by the $\mathcal{I}_{25}$ one. The worst-performing model-centric explanation is that generated from the whole set of important replies ($\mathcal{I}$). We calculate the cumulative importance score of these data partitions and note $\mathcal{I}_{25}$ and $\mathcal{I}_{50}$ contain 75\% and 93\% respectively of the thread's total importance. This suggests the remaining second half of the importance-ordered thread offers little relevant information towards the model's decision. 

\section{Human Evaluation of LLM-based Evaluators}

\begin{table}[!ht]
    \centering
    \scriptsize
    
    \begin{tabular}{lcc}
         \toprule
         Agreement & Informativeness Detection & Veracity Prediction \\
         \toprule
         Ann - Ann & 82$\%$ & 88$\%$ \\
         \midrule
         Ann - ChatGPT & 69$\%$ & 68$\%$\\
         \midrule
         Ann - ChatGPT 0613 & 64$\%$ & 74$\%$\\
         \midrule
         Ann - GPT-4 & 63$\%$ & 80$\%$\\
         \bottomrule
\end{tabular}
    \caption{Pairwise agreement scores for the overlap between the evaluations of the annotators (Ann) and the LLM. The LLMs are: ChatGPT ("gpt-3.5-turbo-0301"), ChatGPT 0613 ("gpt-3.5-turbo-0613") and GPT-4. The evaluations are conducted for two tasks: informativeness detection and veracity prediction.}
    \label{tab:human_eval}
\end{table}
\normalsize

Our human evaluation study has two goals: 1) quantify the evaluation capability of ChatGPT, the LLM employed in our experiments in Sec. \ref{sec:5} to assess automatic explanations and 2) investigate the performance of ChatGPT against more recently-published LLMs. The results are in Table \ref{tab:human_eval}.

We ran a pilot study on 50 threads randomly sampled, such that each fold and each label type is equally represented for a fair evaluation of the LLM performance. We follow a similar evaluation setup to the work of \citep{atanasova-etal-2020-generating-fact}, who study whether their generated summaries provide support to the user in fact checking a claim. We check the LLM-based evaluation of automatic explanations on two tasks: 1. \textbf{Informativeness Detection}, where an Explanation is classified as either informative or uninformative and 2. \textbf{Veracity Prediction}, where an Informative Explanation is assigned true, false or unverified if it helps determine the veracity of the given claim. 

Two Computer Science PhD candidates proficient in English were recruited as annotators for both tasks. Each annotator evaluated the test set of explanation candidates, resulting in 300 evaluations per annotator. The same guidelines included in the prompt from Table \ref{tab:ex_prompt_guidelines} and examples from Appendix \ref{sec:appendixA} are used 
as instructions. Before starting, the research team met with the annotators to ensure the tasks were understood, a process which lends itself to a richer engagement with the guidelines.

\subsection{Evaluation of ChatGPT}

\label{6.1}
\paragraph{Informativeness Detection}
In our first human experiment (Table \ref{tab:human_eval}: first column), we evaluate whether ChatGPT correctly identifies an informative explanation. We find that the agreement between our annotators is 82$\%$ 
which we set as the upper threshold for comparison. We note that the agreement between human evaluators and ChatGPT consistently remains above the random baseline, but experiences a drop. Fleiss Kappa is $\kappa = 0.441$, which is higher than the agreement of $\kappa = 0.269, 0.345, 0.399$ reported by \citet{atanasova-etal-2020-generating-fact} for the same binary setup. After examining the confusion matrix for this task (See Appendix \ref{sec:appendixB}), it is observed that most mismatches arise from false positives - ChatGPT labels an Explanation as informative when it is not. Finally, we find this type of disagreement occurs in instances when the rumour is a complex claim, i.e., a claim with more than one check-worthy piece of information within it. As suggested by \citet{chen-etal-2022-generating}, the analysis of complex real-world claims is a challenging task in the field of fact checking and we also observe its impact on our LLM-based evaluation for rumour verification.

\paragraph{Veracity Prediction}
In our second human experiment (Table \ref{tab:human_eval}: second column), we evaluate if ChatGPT correctly assigns a veracity label to an Informative explanation. Again, we consider 88$\%$, the task annotator agreement to be the upper threshold. Despite the more challenging set-up (ternary classification instead of binary), the LLM maintains good agreement: Fleiss Kappa $\kappa=0.451$ (again higher than those of \citet{atanasova-etal-2020-generating-fact} for the multi-class setup  $\kappa = 0.200, 0.230, 0.333$). Manual inspection of the disagreement cases reveals that the most frequent error type (58 / 75 mislabelled cases exhibit this pattern - See Appendix \ref{sec:appendixB}) is when ChatGPT classifies a rumour as unverified based on the Explanation, while the annotator marks it as true. We hypothesise that an LLM fails to pick up on subtle cues present in the explanation that are otherwise helpful for deriving a veracity assessment. For instance, the Explanation \textit{"I think channel 7 news is saying he [the hostage-taker] is getting agitated bcoz of it [the hostage's escape], its time to go in."} implies that the escape indeed took place as validated by Channel 7; this cue helps the annotator assign a true label to the corresponding claim \textit{"A sixth hostage has escaped from the Lindt cafe in Sydney!"}.

We acknowledge the limitations of using an LLM as an evaluator, which reduces the richness of annotator interaction with the task, but show through our human evaluations that good agreement between an LLM and humans can still be achieved. This not only allows the scaling of final results to the entire dataset instead of being confined to a small test set (See Sec. \ref{Sec:3}), but also provides an automated benchmarking of generated explanations when the ground truth is missing.

\subsection{Comparison to other LLMs}

As ChatGPT is a closed-source tool continually updated by its team, it is important to investigate how ChatGPT-powered evaluations are influenced by the release of newer versions of the same language model or by substitution with improved models. To this effect, we compare the legacy version of ChatGPT released on 1 March 2023 with its more recent version, ChatGPT 0613 (released on 13 June 2023) and finally with GPT-4, a multi-modal model equipped with broader general knowledge and more advanced reasoning capabilities.

We note that that while there are differences between the labels produced by the two versions, there is a higher agreement with human judgement for the newer snapshot ChatGPT 0613 when assessed on the more complex task of veracity prediction. A similar behaviour is observed for GPT-4, whose performance is the most aligned with human judgment in the second task. After examining the error patterns (See Appendix \ref{sec:appendixB}), we observe a notable difference between ChatGPT-based models and GPT-4: while both temporal snapshots of ChatGPT tend to evaluate irrelevant explanations as informative (See Sec. \ref{6.1}), GPT-4 suffers from assigning too many false negatives. This implies the existence of a positive bias for ChatGPT models and a negative bias for GPT-4.

Based on our limited findings, we hypothesise that more recent models have the potential to be more reliable evaluators of explanations than older models, given their higher agreement with human annotators. However, the model choice needs to be grounded into the task requirements (i.e., which errors should be prioritised) and availability of computational costs (at the moment of writing GPT-4 is 20x more expensive than ChatGPT).

\section{Conclusions and Future Work}

We presented a novel zero-shot approach for generating abstractive explanations of model predictions for rumour verification. Our results showed abstractive summaries constructed from important posts scored by a post-hoc explainer algorithm can be successfully used to derive a veracity prediction given a claim and significantly outperform extractive and model-independent baselines. We also found using an LLM-based evaluator for assessing the quality of the generated summaries yields good agreement with human annotators for the tasks of informativeness detection and veracity prediction.

In future work, we plan to jointly train the veracity prediction and explanation generation and assess how an end-to-end approach impacts the quality of resulting explanations. Additionally, we aim to enrich the explanations by incorporating external sources of information such as PHEMEPlus \citep{dougrez-lewis-etal-2022-phemeplus}. Another direction is generating fine-grained explanations for addressing all check-worthy aspects within complex claims.


\section*{Limitations}

\paragraph{Summarisation of threads} 
The format of the conversation threads is challenging to summarise. Our approach to summarisation is to flatten the conversation tree and to concatenate the individual posts, which are then used as an input to a BART model. This approach is na\"{i}ve as the meaning of the nested replies can be lost if considered independently of the context.

\paragraph{Task limitation} 
At the moment, the explanations are constructed exclusively from the information present in the thread. Consequently, the degree of evidence present in a thread is reflected into the explanatory quality of the summary. 

\paragraph{Complex Claims} As seen in the paper, complex claims are a challenging subset of rumours to evaluate. Using the heuristic outlined in \citet{chen-etal-2022-generating} to identify complex claims based on verb count, we find that 22\% of the claims within PHEME are classified as complex. To generate comprehensive explanations covering each check-worthy aspect within such claims, a re-annotation of PHEME is required which is only labelled at claim-level at the moment.

\paragraph{Human Evaluation} Evaluation via large language models is in its infancy. While there have been very encouraging recent results of using it as a viable alternative to human evaluation, these are still early days. It is unclear how much the evaluation stability is impacted by prompt design or by substitution with open-source language models.

\paragraph{Evaluation criteria for generated output} Since our explanations rely on generation mechanisms including automatic summarisers, it is important to acknowledge that there are other evaluation criteria native to the generation field which are outside the scope of this paper and have not been covered. We note that since hallucination, redundancy, coherence and fluency have already been tested in the original works \citep{lewis-etal-2020-bart,bilal-etal-2022-template} introducing the summarisers we employ, we prioritised the criteria relevant to explainable fact-checking in the experiments of this paper: informativeness of explanations and faithfulness to predicted veracity label.

\section*{Ethics Statement}
Our experiments use PHEME dataset, was given ethics approval upon its original release. However, we note that the dataset contains many instances of hate speech that may corrupt the intended aim of the summaries. In particular, summaries that use the majority of posts within the thread may exhibit hate-speech content exhibited by parts of the input text.

\bibliography{anthology,custom}

\appendix

\section{Examples of Assessing the Informativeness of Explanations}
\label{sec:appendixA}

\begin{table*}[!h]
    \centering
    \small
    \begin{tabular}{p{15cm}}
\toprule
\textbf{Claim}: Victims were forced to hold a flag on the cafe window.\\
\textbf{Explanation}: Users believe this is true and point to the released footage.\\
\textbf{Your answer}: A \\ \\

\textbf{Claim}: BREAKING: Hostages are running out of the cafe $\#$sydneysiege\\
\textbf{Explanation}: Some users believe the claim is unverified as Channel 9 did not confirm and some agree that the details of potential escape should not be disclosed.\\
\textbf{Your answer}: C \\ \\

\textbf{Claim}: One of the gunmen left an ID behind in the car.\\
\textbf{Explanation}: One of the gunmen left an ID behind in the car. The majority deny the ID was found there and point to the media for blame.\\
\textbf{Your answer}: B \\ \\

\textbf{Claim}: Three people have died in the shooting.\\
\textbf{Explanation}: Three people have died in the shooting. Most users pray the attack is over soon.\\
\textbf{Your answer}: D \\ \\

\textbf{Claim}: NEWS $\#$Germanwings co-pilot Andreas Lubitz had serious depressive episode (Bild newspaper) $\#$4U9525 URL LINK\\
\textbf{Explanation}:Germanwings co-pilot Andrés Lubitz has serious depressive episode. Never trust bild. Users believe that bild is a fake newspaper and the stories concerned with the suicide of Andreas Lubitz should not be discussed.\\
\textbf{Your answer}: C \\ \\

\textbf{Claim}: Snipers set up on National Art Gallery as we remain barricaded in Centre Block on Parliament Hill $\#$cdnpoli.\\
\textbf{Explanation}: Snipers set up on National Art Gallery as we remain barricaded in Centre Block on Parliament Hill. Most users are skeptical about the news and await more details.\\
\textbf{Your answer}: C \\ \\

\textbf{Claim}: BREAKING: $\#$Germanwings co-pilot's name is Andreas Lubitz, a German national, says Marseilles prosecutor. \\
\textbf{Explanation}: He didn’t have a political or religious background.\\
\textbf{Your answer}: D \\ \\

\textbf{Claim}: Several bombs have been placed in the city \\
Explanation: This is false, why then cause panic and circulate on social media? \\ 
\textbf{Your answer}: B \\ \\

\textbf{Claim}: Police report the threats released by the criminals. \\
\textbf{Explanation}: The majority threaten to condemn anyone who is a terrorist. \\
\textbf{Your answer}: D \\ \\

\textbf{Claim}: $\#$CharlieHebdo attackers shouted 'The Prophet is avenged'. \\
\textbf{Explanation}: In video showing assassination of officer.walking back to car they shouted: 'we avenged the prophet.We killed Charlie Hebdo' \\
\textbf{Your answer}: A \\

\bottomrule
          
    \end{tabular}
    \caption{Ten representative examples covering diverse explanation styles and veracity labels are selected. These are included in the final prompt for ChatGPT.}
    \label{tab:ex_prompt}
\end{table*}

\newpage

\section{Error Analysis of LLM's performance as Evaluator}
\label{sec:appendixB}
We note that our ChatGPT-human agreement scores for both tasks are similar or higher to those reported by \citet{pheme_original}, who employ crowd-sourced workers for annotating similar classification subtasks on the PHEME dataset: 61.1$\%$ for labelling certainty of rumours and 60.8$\%$ for classifying types of evidence arising from the thread.

We report the performance of ChatGPT, ChatGPT 0614 and GPT-4 as evaluators using the manually annotated set of 200 explanations. The error analysis is shared via a confusion matrix for each task: informativeness detection (See Table \ref{tab:cf1}) and veracity prediction (See Table \ref{tab:cf2}). The results are reported as counts.

\begin{table}[!ht]
    \centering
    \footnotesize
    
    \begin{tabular}{ccc}
         \toprule
          \diagbox{LLM}{Annotator} & Informative & Uninformative \\
          \midrule
         & ChatGPT & \\
         \midrule
         Informative & 169 & 107 \\
         Uninformative & 81 & 143 \\

         \midrule
         & ChatGPT 0613 &\\
         \midrule
         Informative & 236  & 104 \\
         Uninformative & 114  & 146  \\

        \midrule
         & GPT-4 & \\
         \midrule
         Informative & 160 & 30 \\
         Uninformative & 190 & 220 \\
         \bottomrule
                
    \end{tabular}
    \caption{Confusion Matrices for ChatGPT, ChatGPT 0613 and ChatGPT-4 for the task of \textbf{Informativeness Detection}}
    \label{tab:cf1}
\end{table}

\begin{table}[!ht]
    \centering
    \footnotesize
    \begin{tabular}{cccc}
         \toprule
          \diagbox{LLM}{Annotator} & True & False & Unverified \\
          \midrule
         & & ChatGPT & \\
         \midrule
         True & 105 & 3 & 4\\
         False & 12 & 18 & 5 \\
         Unverified & 58 & 3 & 61 \\
         \midrule
               
        & & ChatGPT 0613 & \\
        \midrule
         True & 114  & 3 & 8 \\
         False & 10 & 10  & 6 \\
         Unverified & 26 & 8 & 51 \\
         \midrule

        & & GPT-4 & \\
        \midrule
         True & 78  & 0 & 2 \\
         False & 10 & 10  & 9 \\
         Unverified & 7 & 84 & 40 \\
         \bottomrule
                
    \end{tabular}
    \caption{Confusion Matrices for ChatGPT, ChatGPT 0613 and ChatGPT-4 for the task of \textbf{Veracity Prediction}}
    \label{tab:cf2}
\end{table}
\normalsize

\section{Pilot Study on Temperature Setting for ChatGPT}
\label{sec:appendixC}

We used the same explanations in Table 4 and ran a small pilot study to assess how incrementing the temperature parameter affects the LLM evaluation. Results are in Table \ref{tab:temperature}. We used increments of 0.2 in temperature and ran the experiment 3 times to account for the non-deterministic behaviour. Overall, the evaluations remain consistent (94$\%$ of the labels output by ChatGPT are the same) across runs and temperature values. In particular, we note that when using temperature 0, the evaluations remain 100$\%$ consistent and for non-zero temperature, the evaluation only impacts the labelling of the last explanation which is less helpful than previous explanation candidates.

\begin{table*}[]
    \centering
    \small
    \begin{tabular}{p{6cm}cccccc}
\textbf{Explanation} & $T=0$ & $T=0.2$ & $T=0.4$ & $T=0.6$ & $T=0.8$ & $T=1$\\
\midrule
@TorontoStar Ok, time to take it to the ***muslims. Look out Allah, here comes the revenge. ***. & D,D,D& D,D,D& D,D,D& D,D,D& D,D,D& D,D,D\\
\\
Soldier dies from shooting in Ottawa and Parliament Hill is in lockdown. The majority think the media is wrong to report that Parliament Hill was in lockdown and that the lockdown was a ploy to target Muslims. &B,B,B& B,B,B& B,B,B& B,B,B& B,B,B& B,B,B\\ \\
Cdn soldier dies from shooting dead in Ottawa. The majority are sceptical about the news of the shooting and some are questioning where the confirmation is coming from.& C,C,C& C,C,C& C,C,C& C,C,C& C,C,C& C,C,C\\ \\
Cdn soldier dies from shooting in Ottawa and Parliament Hill is in lockdown. Most users ask where the news of the gunman is and are wondering who is responsible for his death. Many of the responses use humour and irony, such as: ’I don’t think the soldier is dead’.& C,C,C& C,A,C& C,C,C& C,C,C& C,A,A& C,C,A\\

    \end{tabular}
    \caption{Labels output by ChatGPT for each explanations across 3 different runs.}
    \label{tab:temperature}
\end{table*}
\normalsize

\section{Experimental Setup}
\label{sec:appendixD}

We train the rumour verification model for 300 epochs with learning rate $10^{-5}$. The training loss is cross-entropy. The optimizer algorithm is Adam \citep{adam}. Hidden channel size is set as 256 for the propagation and dispersion components and 32 hidden channel size for the stance component. The batch size is 20. 
For the GraphSage layers, we apply a mean aggreggator scheme, followed by a relu activation. For the Multi-headed Attention layer, we use 8 heads.
Embeddings generated by the "all-MiniLM-L6-v2" model from Sentence Transformers \citep{reimers-2019-sentence-bert} are used to initialise the node representations in the graphs.
To avoid overfitting, we randomly dropout an edge in the graph networks with probability 0.1. We use a Nvidia A5000 GPU for our model training. All model implementation is done via the \textit{pytorch-geometric} package \citep{Fey/Lenssen/2019} for graph neural networks.



\end{document}